\pdfoutput=1

\documentclass[11pt]{article}

\usepackage[preprint]{acl}

\usepackage{times}
\usepackage{latexsym}

\usepackage[T1]{fontenc}

\usepackage[utf8]{inputenc}

\usepackage{microtype}

\usepackage{inconsolata}

\usepackage{graphicx}

\usepackage{soul}

\usepackage{graphicx} 
\usepackage{makecell}
\usepackage{amsmath}

\usepackage{colortbl}
\usepackage{array}

\usepackage{xcolor}
\usepackage{xspace}

\usepackage{multirow}
\usepackage{subfigure}

\usepackage{booktabs}
\usepackage{amsfonts}

\newcommand{\method}{ImageRef-VL\xspace}

\title{\method: Enabling Contextual Image Referencing \\ in Vision-Language Models}

\author{
  Jingwei Yi$^{1*}$, Junhao Yin$^{2}$, Ju Xu$^{2}$, Peng Bao$^3$, Yongliang Wang$^2$, Wei Fan$^{4}$, Hao Wang$^{5\dag}$ \\
  $^1$University of Science and Technology of China 
  $^2$ByteDance \\
  $^3$Peking University 
  $^4$University of Oxford 
  $^5$Tsinghua University \\
  {\tt yjw1029@mail.ustc.edu.cn} {\tt weifan.oxford@gmail.com}
  {\tt hao-wang20@mails.tsinghua.edu.cn} \\
  {\tt \{yinjunhao,yufeng.1016,baopeng.peter,yongliang.wyl\}@bytedance.com} \\
}

\begin{document}
\maketitle

\def\thefootnote{*}\footnotetext{This work was done when the author Jingwei Yi was at Bytedance Group for intern.}\def\thefootnote{\arabic{footnote}}
\def\thefootnote{\dag}\footnotetext{Corresponding authors.}\def\thefootnote{\arabic{footnote}}

\begin{abstract}

Vision-Language Models (VLMs) have demonstrated remarkable capabilities in understanding multimodal inputs and have been widely integrated into Retrieval-Augmented Generation (RAG) based conversational systems. 
While current VLM-powered chatbots can provide textual source references in their responses, they exhibit significant limitations in referencing contextually relevant images during conversations.
In this paper, we introduce \textit{Contextual Image Reference} -- the ability to appropriately reference relevant images from retrieval documents based on conversation context -- and systematically investigate VLMs' capability in this aspect.
We conduct the first evaluation for contextual image referencing, comprising a dedicated testing dataset and evaluation metrics. 
Furthermore, we propose \method, a method that significantly enhances open-source VLMs' image referencing capabilities through instruction fine-tuning on a large-scale, manually curated multimodal conversation dataset.
Experimental results demonstrate that \method not only outperforms proprietary models but also achieves an 88\% performance improvement over state-of-the-art open-source VLMs in contextual image referencing tasks. 
Our code is available at \url{https://github.com/bytedance/ImageRef-VL}.

\end{abstract}    
\section{Introduction}
\label{sec:intro}


\begin{figure}[!t]
  \centering
  \includegraphics[width=0.45\textwidth]{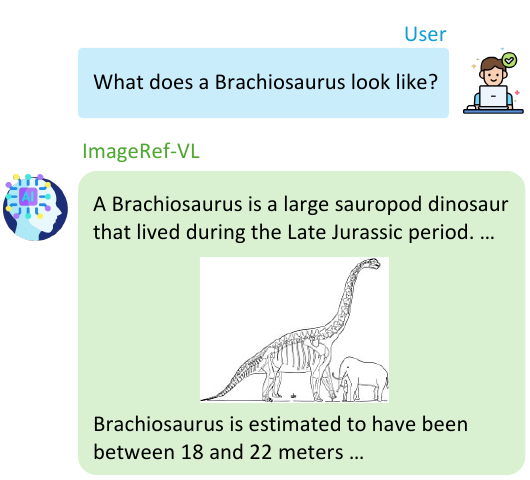}
  \caption{An example of contextual image reference, where referencing the images of a Brachiosaurus can largely enhance user comprehension and engagement.\protect\footnotemark}
  \label{fig:example}
\end{figure}
\footnotetext{The response was generated with reference to \url{https://en.wikipedia.org/wiki/Brachiosaurus.}}

In recent years, Vision-Language Models (VLMs) have achieved remarkable progress, enabling advanced multi-modal reasoning and generation from combined text and image inputs.
Both close-source models, such as GPT-4o~\cite{hurst2024gpt} and Claude~\cite{link_claude}, and open-source models, such as Phi-3.5-Vision~\cite{abdin2024phi}, Qwen2-VL~\cite{wang2024qwen2} and InternVL2~\cite{chen2024internvl}, have demonstrated impressive capabilities across a range of vision-language tasks~\cite{hudson2019gqa,yu2023mm,fu2024mmecomprehensiveevaluationbenchmark}. 
Concurrently, the integration of VLMs with Retrieval-Augmented Generation (RAG) has allowed these models to retrieve and incorporate external knowledge into their responses\footnote{\url{https://chatgpt.com/}}\footnote{\url{https://claude.ai/}}\footnote{\url{https://www.doubao.com/chat/}}\footnote{\url{https://www.perplexity.ai/}}. 
One feature of VLM-based RAG chatbots is to provide references to the retrieved text sources used to generate responses.
Extensive research efforts have been dedicated to improving textual reference accuracy in RAG chatbots~\cite{gao2023enabling,shen2024citekit,zhang2024longcite,fierro2024learning}.

While VLM-based RAG systems have made significant strides in providing reliable textual references, they face a critical limitation in their ability to leverage visual content effectively during conversations.
We identify this gap as the absence of \textit{Contextual Image Reference} - the capability to strategically select and incorporate relevant images from retrieved documents to enhance response comprehension and user engagement.
As demonstrated in Figure~\ref{fig:example}, when discussing complex subjects like the Brachiosaurus, purely textual descriptions of its physical characteristics often fail to convey information intuitively. 
Despite its potential impact on multimodal conversation systems, the challenge of contextual image referencing remains largely unexplored in current research.

To address these challenges, we first propose and formally define \textit{Contextual Image Reference} as a novel task that requires VLMs to incorporate relevant images as integral components of their responses.
To systematically evaluate performance on this task, we construct a dedicated testing dataset and develop comprehensive metrics that assess a model's capability to integrate images in contextually appropriate ways.
Building upon open-source VLMs, (i.e., InternVL2), we present \method, a framework that enhances models' contextual image referencing abilities.
Our approach involves generating initial responses using existing LLMs and VLMs, carefully curating these outputs through manual refinement, and leveraging the resulting high-quality dataset for supervised fine-tuning.
Through this process, \method learns to make informed decisions about when and how to incorporate images as authoritative visual references.

The primary contributions of our work are summarized as follows:

\begin{itemize}
    \item To the best of our knowledge, we are the first to introduce and formally define \textit{Contextual Image Reference} as a novel task for multimodal conversational AI, addressing a critical gap in current VLM capabilities.
    \item We conduct a comprehensive evaluation for this task, including a carefully curated testing dataset and novel metrics that capture both the relevance and naturalness of image references.
    \item We propose \method, a fine-tuning framework that significantly advances the state-of-the-art in contextual image referencing, demonstrating an 88\% performance improvement over existing open-source VLMs.
    \item Through extensive experiments across various scenarios, we validate the effectiveness of our approach and establish strong baseline results for future research.
\end{itemize}
\section{Related Works}
\label{sec:related-work}


\subsection{Vision Language Models}
Large Language Models (LLMs), primarily based on the transformer architecture~\cite{vaswani2017attention}, have recently achieved remarkable performance in a wide range of natural language tasks~\cite{zhang2023prompting,poesia2022synchromesh,kojima2022large}. 
Building upon these advances, Vision-Language Models (VLMs) extend LLM capabilities to the visual domain, enabling sophisticated reasoning and content generation from both textual and visual inputs~\cite{link_llama_32,lu2024deepseek,liu2024visual,liu2024improved}.
Recently, numerous VLMs have been introduced, spanning both close-souced systems (e.g., GPT-4o~\cite{hurst2024gpt}, Claude~\cite{link_claude}) and open-source frameworks (e.g., Phi-3.5-Vision~\cite{abdin2024phi}, Qwen2-VL~\cite{wang2024qwen2}, InternVL2~\cite{chen2024internvl}). These models have demonstrated impressive capabilities across diverse vision-language benchmarks~\cite{hudson2019gqa,yu2023mm,fu2024mmecomprehensiveevaluationbenchmark}, promoting an evolving research landscape in multimodal machine intelligence.
Existing VLMs are generally composed of three core components, i.e., a vision encoder, an adapter, and an LLM back-end.
The vision encoder, such as CLIP~\cite{radford2021learning} or BLIP~\cite{li2022blip,li2023blip}, is designed to extract rich visual features from images, converting them into representations that can be effectively processed by downstream components.
The adapter component subsequently bridges these extracted features to the language model, employing architectures such as simple multi-layer perceptrons~\cite{liu2024visual,liu2024improved} or  Q-formers~\cite{li2023blip} 
Finally, the LLM generates responses by combining the visual and textual inputs.


\subsection{Retrieval-Augmented Generation}
Retrieval-Augmented Generation (RAG) is a technique designed to enhance the capabilities of LLMs by integrating external knowledge sources~\cite{gao2023retrieval,gupta2024comprehensivesurveyretrievalaugmentedgeneration}.
It addresses key challenges of LLMs, such as hallucinations~\cite{huang2023survey,tonmoy2024comprehensive,xu2024hallucination,fan2024survey} and time-sensitive information~\cite{mousavi2024your}, by incorporating relevant information retrieved from external databases or documents. 
The RAG process typically involves three main steps: retrieval, generation, and augmentation.
First, a retriever identifies and extracts relevant document chunks from a knowledge base based on semantic similarity to the user's query. 
These retrieved chunks are then combined with the original query to form an augmented context, which is used as input for the LLM to generate a response. 

\subsection{LLM Citation Generation}

LLM Citation Generation has gained attention as a way to enhance the verifiability and transparency of model-generated responses by including citations linked to external evidence. 
Citation generation improves the factual accuracy of LLMs answers~\cite{gao2023enabling} and allows users to trace the sources of information, thereby increasing the credibility and explainability of outputs~\cite{tahaei2024efficient}. 
Early work like ALCE~\cite{gao2023enabling} proposed foundational methods and evaluation metrics for enabling LLMs to generate citations. 
Subsequent studies have improved citation quality through fine-tuning approaches~\cite{huang2024training,li2024improving,ye2024effective} or multi-stage pipelines~\cite{zhang2024verifiable,hennigen2023towards,lee2023towards}.
Although existing works have explored the citation generation for LLMs, no prior studies have addressed the problem of contextual image reference in multimodal settings or proposed corresponding solutions.

\section{Problem Definition.}
In this section, we present the problem definition for contextual image reference.
Given an input prompt consisting of an ordered sequence of mixed images and text, denoted as $\{E_1, E_2, \dots, E_m\}$, where each element $E_i$ is either an image $I_i$ from the set $I = \{I_1, I_2, \dots, I_n\}$ or a text segment $T_i$ from the set $T = \{T_1, T_2, \dots, T_k\}$, the model's goal is to generate a textural response $R$ that meets the prompt's requirements.
The response $R$ can be with some images referenced through a contextual image ID that falls within the range $[1, n]$, corresponding to the images in the input set. These image references must be contextually appropriate and align with the textual descriptions or contextual requirements specified in the prompt, ensuring the response is coherent, relevant, and adheres to the prescribed format.

\begin{figure*}[!t]
  \centering
  \includegraphics[width=0.95\textwidth]{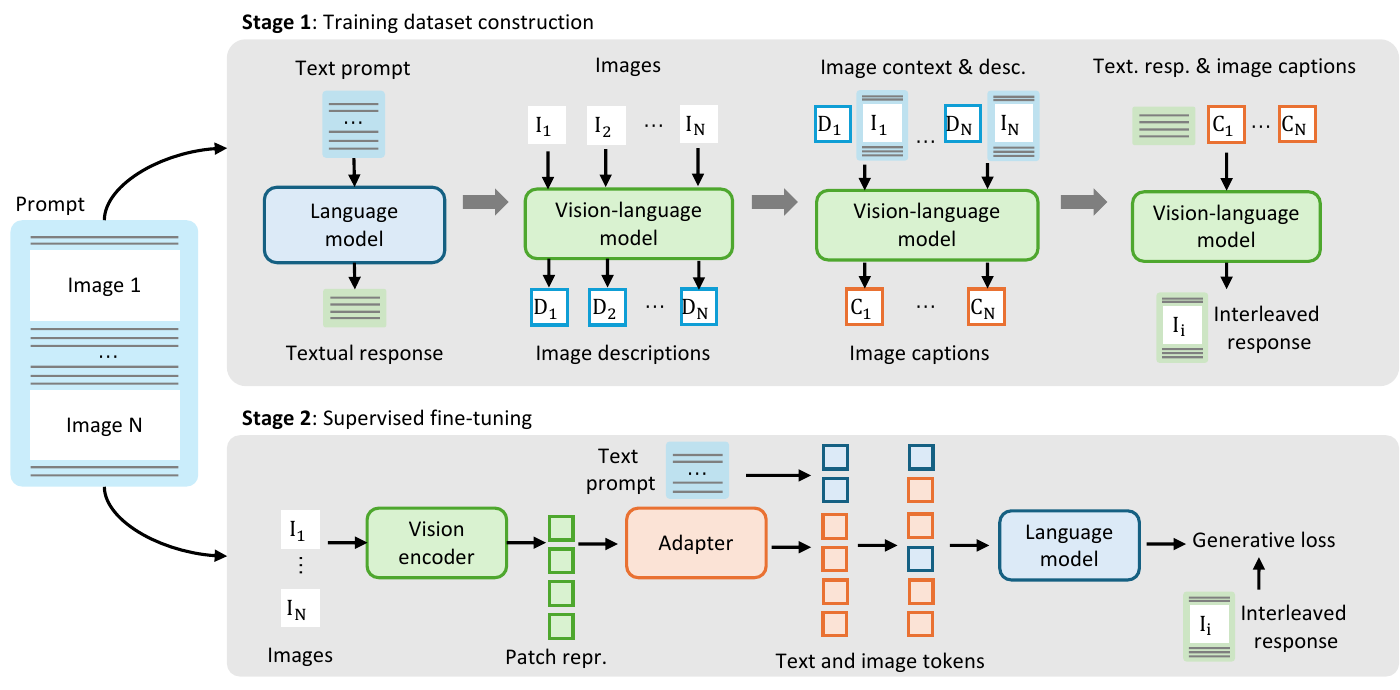}
  \caption{\textbf{The training strategy of the proposed IMI-VL model}.  Stage 1: Training dataset construction involves generating textual responses and image descriptions through a language model and a vision-language model. These are combined into interleaved responses using image contexts and captions. Stage 2: Supervised fine-tuning refines the model with a vision encoder, adapter, and language model, optimizing through generative loss.}
  \label{fig:method}
\end{figure*}

\section{Method}
To enhance the contextual image referencing capability of vision-language models, it is critical to incorporate contextually relevant image-text data during the model's training phase. Specifically, datasets that feature contextually integrated image references should be included in the supervised fine-tuning (SFT) stages of the model's development. 
We construct the training dataset by incorporating image references from the retrieved documents into appropriate positions within the original text-only responses generated by the LLM.
To achieve this goal, two challenges need to be addressed. The first is to collect an SFT dataset containing responses with contextual image references. The second is to effectively fine-tune existing VLMs.

\noindent \textbf{Overview.}
To address the initial challenge, we developed a method to generate responses with contextual image references using existing LLMs, VLMs and user prompts. Given a prompt, we first generate a text-based response, then create a caption for the input image using the VLM, and integrate this caption into the text. Our model training follows the VLM’s standard SFT approach, but we enhance image understanding by requiring the model to refer explicitly to input images.
To preserve the general VLM capabilities, we combine the original SFT data with interleaved multi-image SFT data. The \method framework is illustrated in Figure~\ref{fig:method}.

\subsection{Training Data Construction}
\label{sec:data-construct}
To construct the training dataset, we use existing LLMs and VLMs to create a high-quality dataset through a multi-stage few-shot learning approach, and then manually select qualified samples for model training. This method significantly reduces the labeling effort. Specifically, the data generation process involves three steps: generating text-based responses from pure text content, generating captions for each image based on text context, and adding the images references into the responses.

\noindent \textbf{Text Response Generation.}
In this step, we remove the images from the prompt and provide some reference text for the model to generate a pure text response.

\noindent \textbf{Image Caption Generation.}
In this step, we need to generate context-based image descriptions. The image description should not only include information about the image itself but also complement the information related to the image described in the text. For example, when describing an image in the Wikipedia page about Einstein, it is not enough to simply say, `This is a portrait of an elderly man with white hair'; we also need to complete the information by adding, `This is a portrait of Einstein in his later years.' However, directly using the existing VLM to perform this task can lead to over-reliance on context: when the image is unrelated to the context, the model may incorrectly force a connection, and some contextual information may be left incomplete.

To address this issue, we propose a two-stage image description generation approach. In the first stage, we generate a description based solely on the image itself. In the second stage, we provide the image description along with the context to the VLM, asking it to supplement the missing information. We use in-context learning and include examples in both stages to guide the VLM on how to better generate the image description and complete the information.

\noindent \textbf{Image Insertions.}
We input the generated image captions and text responses into an LLM, asking the model to insert the images into the text response. The final results, after manual filtering, form our IMI-interleave training dataset.

\noindent \textbf{Mixture of Datasets.}
To ensure the LLM truly understands the images in context, we also incorporate a certain proportion of contextual image caption generation tasks in the training set, where VLMs generate captions for multiple images in the prompt based on the context, with the captions being those produced in the second step. Additionally, to prevent a significant decline in the performance of VLMs on other image-related tasks, we mix in a proportion of the InternVL2 SFT dataset into our training set.

\subsection{Model Training}

We used the constructed dataset to perform supervised fine-tuning on the VLM to enhance its ability to understand interleaved multi-image tasks. In this subsection, we will briefly introduce the model architecture and training loss.

\noindent \textbf{Model Architecture.}
As shown in stage 2 of Figure~\ref{fig:method}, our fine-tuned model consists of three components: the vision encoder, the adapter, and the language model.
Given a user prompt and retrieved documents with images, the vision encoder processes each image, extracting patch features from the image. The adapter, acting as a bridge between the vision encoder and the language model, maps the patch features into the embedding space of the language model, resulting in image tokens. The textual part of the prompt is tokenized and embedded to obtain text tokens. Finally, the text tokens and image tokens are fed into the language model in their original positions for modeling.


\noindent \textbf{Training Loss.}
We proceed the typical supervised fine-tuning process~\cite{ouyang2022training}. The dataset $\mathcal{D}$ is composed of $N$ prompt-response pairs: $\mathcal{D}=\{ (\textbf{x}^i, \textbf{y}^i) \}_{i=1}^N$, where both prompt $\textbf{x}^i$ and ground-truth response $\textbf{y}^i$ are a sequence of tokens. We denote $p(\textbf{y}^i_j | \textbf{x}^i \oplus \textbf{y}^i_{<j})$ as the probability of the outputting next token as $\textbf{y}^i_j$ given previous tokens $\textbf{x}^i \oplus \textbf{y}^i_{<j}$, where $\oplus$ is the concatenation operator and $\textbf{y}^i_{<j}$ denotes the tokens before index $j$. The training loss is then $\mathcal{L}=-\log \prod_{j=1}^{n^i} p(\textbf{y}^i_j | \textbf{x}^i \oplus \textbf{y}^i_{<j})$, with $n^i$ being the length of $\textbf{y}^i$ and the optimization variable being a VLM.

\section{Evaluation Settings.}
\label{sec:eval}
To evaluate the performance of model supporting contextual image references, we propose three kinds of evaluation metrics, i.e., textual content evaluation, image position evaluation, overall response evaluation.

\subsection{Automated Evaluation}

\noindent \textbf{Text Evaluation.}
When a model generates responses with contextual image references, we can evaluate the textual portion of the response. Following the existing LLM-as-judge approach~\cite{zheng2023judging}, allowing a large language model to score the response and then calculating the candidate model's average score across all test samples. Considering that current large language models may not perform well in understanding multi-image prompts, we only include the textual prompt and provide a reference answer for scoring.

\noindent \textbf{Image Position Evaluation.}
Inspired by existing work on image position prediction~\cite{muraoka2020image}, when a model generates contextual image references, we can evaluate whether each image is placed in an appropriate position. However, previous approaches rely on an existing image-inserted dataset, checking whether the model places the image in a specific, pre-defined position or if the image ranks within the top-K choices. However, this metric does not align well with the actual user experience, as multiple images within the candidate pool could be suitable for position $i$. 

To address this issue, we redesigned the image position evaluation metric. Specifically, for each potential image insertion point, we classify all images into four categories:
\begin{itemize}
    \item 3-point images: Images that perfectly match the current contextual content.
    \item 2-point images: Images that match the current contextual content but are of low quality (e.g., blurry, obstructed).
    \item 1-point images: Images related to the main subject mentioned in the current context.
    \item  0-point images: Images not related to the current contextual content.
\end{itemize}
Finally, we can obtain a testing dataset in the following format: 

\begin{equation}
\begin{aligned}
\mathcal{D}_{\text{test}} = \Big\{ \, & p_i : \Big\{ s : \big[ I_{sp_i}^1, \dots, I_{sp_i}^{M_{sp_i}} \big] \, \Big| \\
& \, s \in [0, 3] \Big\} \,  \Big| \,  p_i \in P_i, \, i \in \left[ 1, N \right] \Big\},
\end{aligned}
\end{equation}
where $P_i$ is all potential image insertion points of the $i$-th testing sample, $I_{sp_i}^j$ is the $j$-th image with $s$ point for position $p_i$, $M_{sp_i}$ is the maximum number of $s$-point images at position $p_i$, and $N$ is the number of samples in the testing dataset.

Based on this label definition, we designed three metrics: Precision, Recall3, F1 and Score.
\begin{itemize}
    \item \textbf{Precision} is defined as the accuracy of the illustrations, meaning the probability that an inserted image is a nonzero score image. It is defined as follows:
\begin{equation}
\text{Precision} = \frac{\sum_{i=1}^{N} \mathbb{I}(s_i > 0)}{N},
\end{equation}
where $N$ is the total number of inserted images, and $\mathbb{I}(s_i > 0)$ is an indicator function that returns 1 if the score $s_i$ of the $i$-th image is greater than zero, and 0 otherwise.
    \item \textbf{Recall3} is defined as the coverage rate of the relevant images, representing the probability that the images were inserted at all possible positions where the 3-point images could be inserted. Since the same image cannot be inserted in two different positions, we used the BPM algorithm to calculate the maximum number of relevant images that can be inserted under the current sample's score label. 
    The Recall metric is defined as follows:
\begin{equation}
\text{Recall3} = \frac{\sum_{p \in P} \mathbb{I}(s_p = 3)}{\sum_{p \in P} M_{3p}},
\end{equation}
where $P$ is the set of all possible image insertion points, and $\mathbb{I}(s_p = 1)$ is an indicator function that returns 1 if an image with score $3$ is inserted at position $p$, and 0 otherwise.
    \item \textbf{F1} is the harmonic mean of precision and recall3, which is formulated as follows:
\begin{equation}
\text{F1} = 2 \times \frac{\text{Precision} \times \text{Recall3}}{\text{Precision} + \text{Recall3}}
\end{equation} 

\end{itemize}

\subsection{Human Evaluation}
In addition to evaluating the text generation content and image placement, it is also necessary to assess the overall quality of the generated interleaved response. However, there is currently no effective automated method to evaluate the quality of a text-image interleaved response. Therefore, we employ human evaluation, assigning separate scores for the text, image, and overall response quality. For scoring in all three aspects, we use a 5-point Likert scale~\cite{joshi2015likert} as the rating option.
\section{Experimental Results}

\begin{table}[!t]
\centering
\caption{Statistic details of datasets used in our experiments.}
\scalebox{0.78}{
\begin{tabular}{l|ccc}
\Xhline{1.0pt}
\multicolumn{4}{c}{Train}                                \\ \Xhline{1.0pt}
Dataset   & \# Sample & \# Image & Avg. prompt len.     \\ \midrule
CIR-Interleave  &  7,645  & 73,833    & 5,481.67          \\
CIR-Caption   &   5,633    & 29,558     & 2,215.84          \\
InternVL2-SFT &  1,267,819   & 1,227,131  &  501.54       \\
\Xhline{1.0pt}
\multicolumn{4}{c}{Test}                                 \\
\Xhline{1.0pt}
Dataset   & \# Sample & \# Image & Avg. prompt len.     \\ \midrule
CIR-Test      &   456     & 3,767      & 5,884.99           \\
\Xhline{1.0pt}
\end{tabular}
}
\label{tab:stat}
\end{table}

\begin{table*}[!t]
\centering
\caption{Performance comparison of our \method with baselines. The top method for each metric is in \textbf{bold}, and the second-best is \ul{underlined}.}
\scalebox{0.85}{
\begin{tabular}{l|l|c|ccc|ccc}
\Xhline{1.0pt}
                         \multirow{2}{*}{Method}           &   \multirow{2}{*}{Model}  & Text eval. & \multicolumn{3}{c|}{Image position evaluation} & \multicolumn{3}{c}{Human evaluation} \\ \cmidrule(lr){3-3} \cmidrule(lr){4-6} \cmidrule(lr){7-9}
                                                             &           & Score       & Precision      & Recall3 & F1       & Text & Image & Overall   \\ \midrule
\multirow{3}{*}{Close-sourced VLMs}                          & GPT-4o    &   7.27 &  \cellcolor{gray!20}---    &  \cellcolor{gray!20}---  &  \cellcolor{gray!20}---  &   2.61    &    3.78   &  3.17 \\
                                                             & Claude-3.5-sonnet    & \ul{7.35}   &  \cellcolor{gray!20}---  & \cellcolor{gray!20}---          & \cellcolor{gray!20}---    &   2.60   &   3.91   &   3.32 \\ \midrule
\multirow{3}{*}{Open-sourced VLMs}                          & Phi-3.5-Vision &  2.51   &     \textbf{100.00} & 5.73 & 10.83    &   1.55  &    2.20  &   1.57 \\
                                                            & Qwen2-VL-7B   & 1.95 & \textbf{100.00} & 5.73 &  10.83        &   1.82   &   2.15   & 1.73    \\
                                                             & InternVL2-26B &  4.84   & 79.59 &  10.56 & 18.65  &   1.92  &     2.51    &    1.87    \\  \midrule
\multirow{2}{*}{Three-stage generation}                      & GPT-4o    &  \textbf{7.86} &  66.09 &  \textbf{62.94} & \textbf{64.48}   &      2.77  &   3.80 &   3.25  \\
                                                             & InternVL2-26B & 5.85 & 59.05 & \ul{59.43} & \ul{59.24}  &   2.09   &  3.21  &  2.29  \\ \midrule
\multirow{2}{*}{Our method}                         & \method-8B    & 7.09 & 65.20  & 37.25 &  47.41    &   \textbf{2.97}  &   \ul{4.05}   &  \textbf{3.52} \\
& \method-26B    &    7.30  &    63.75   &   29.26   &  40.11   &   \ul{2.90}  &   \textbf{4.08}    &  \ul{3.36} \\
\Xhline{1.0pt}
\end{tabular}
}
\label{tab:perform}
\end{table*}

\subsection{Experimental Settings}
According to the description in Section~\ref{sec:data-construct}, we constructed two datasets, the CIR-Interleave and CIR-Caption datasets, to enhance the model's ability of contextural image referencing. The CIR-Interleave dataset consists of multiple user prompt and retrieved documents, and the model is required to generate responses with contextural image references based on these texts and user prompts. The CIR-Caption dataset contains contextual caption texts, and the model is tasked with providing contextually relevant captions for each image mentioned within the text. Additionally, we blend the InternVL SFT dataset at a certain ratio to fine-tune the model.
During the testing phase, we construct the CIR-Test dataset as outlined in Section~\ref{sec:eval}, and report evaluation scores on this dataset, including text evaluation scores, image position evaluation metrics such as precision, recall, F1, and human evaluation scores.
Detailed statistics for all the datasets used during training and testing can be found in Table~\ref{tab:stat}.

We conduct experiments on InternVL2-8B and InternVL2-26B~\cite{chen2024internvl}, using 16 A100 GPUs, optimizing the full-fine-tuning training process with DeepSpeed Zero-3 and gradient checkpointing for improved memory efficiency and scalability. Both models are trained with a global batch size of 128, utilizing a learning rate of 4e-5 for the 8B model and 2e-5 for the 26B model, with corresponding weight decay values of 0.01 and 0.05. Training is conducted over 1000 steps, with a maximum sequence length of 16,384 tokens. For \method-26B, we mix the InternVL2-SFT dataset with the combined MI-Interleave and IMI-Caption datasets at a 1:1 ratio and train the model for 550 steps. For \method-8B, we mix the InternVL2-SFT dataset with the combined MI-Interleave and IMI-Caption datasets at a 1:4 ratio and train the model for 950 steps.

\begin{figure*}[!t]
  \centering
  \includegraphics[width=0.95\textwidth]{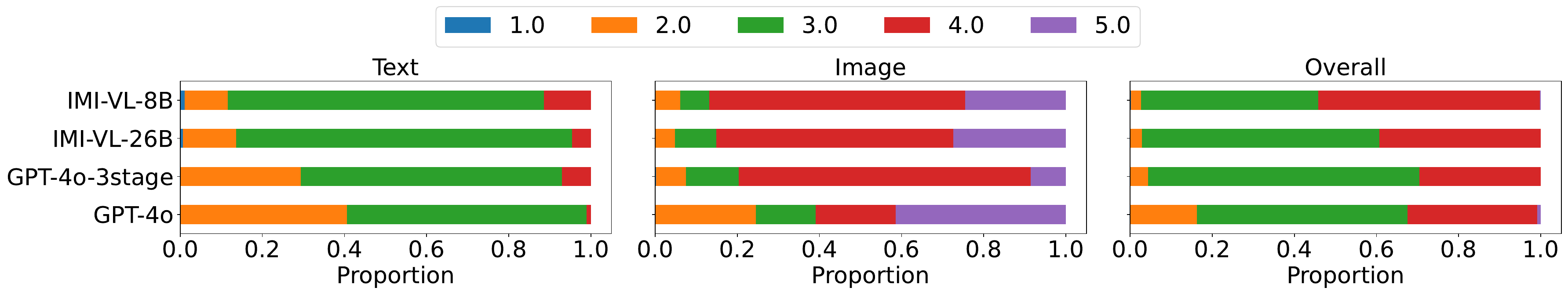}
  \caption{Human evaluation score distribution of four methods.}
  \label{fig:score_dist}
\end{figure*}

\subsection{Performance Comparison}
In the subsection, we compare the performance of \method with the baseline methods, including existing close-sourced vision-language models:
\begin{itemize}
    \item \textbf{GPT-4o}~\cite{hurst2024gpt}: a fast, cost-effective, multimodal large language model by OpenAI.
    \item \textbf{Claude-3.5-Sonnet}~\cite{link_claude}: A multimodal large language model by Anthropic, offering advanced safety and language capabilities.
\end{itemize}
and open-sourced vision-language models:
\begin{itemize}
    \item \textbf{Phi-3.5-Vision}~\cite{abdin2024phi}: a lightweight, open multimodal large language model by Microsoft.
    \item \textbf{Qwen2-VL-7B}~\cite{wang2024qwen2}: the latest version of the vision language models in the Qwen model families.
    \item \textbf{InternVL2-26B}~\cite{chen2024internvl}: the latest addition to the InternVL series of multimodal large language models.
\end{itemize}
and the three-stage response generation method introduced in Section~\ref{sec:data-construct} with GPT-4o and InternVL2-26B.
We report the text evaluation score and human evaluation score for all models. For image position evaluation score, controlled sampling\footnote{https://github.com/dottxt-ai/outlines} is used for open-sourced VLMs to complete image references based on the given text. For three-stage methods, we provide text in prompts and ask the model to insert image references (see Section~\ref{sec:data-construct} Image Insertions part).  Since closed-source VLMs lack controlled sampling capabilities, their image position scores cannot be reported.
The experimental results are shown in Table~\ref{tab:perform}. We also present the detailed distribution of human evaluation scores for GPT-4o, GPT-4o three-stage, as well as our \method-8B and -32B in Figure~\ref{fig:score_dist}.

\noindent \textbf{Effectiveness of \method.} In terms of the human evaluation metrics, \method-8B and \method-26B achieved the best performance in text scores, illustration scores, and overall experience.
An 88\% performance improvement is achieved over state-of-the-art open-source VLMs. 
This demonstrates the effectiveness of our approach.
By further examining Figure~\ref{fig:score_dist}, we can observe that the proportion of severe bad cases produced by \method is significantly lower than that of other methods, with the bad case rate of \method-26B being lower than that of \method-8B.

\begin{figure*}[!t] 
  \centering
  \subfigure[Data mixture ratio.]{\includegraphics[width=0.47\textwidth]{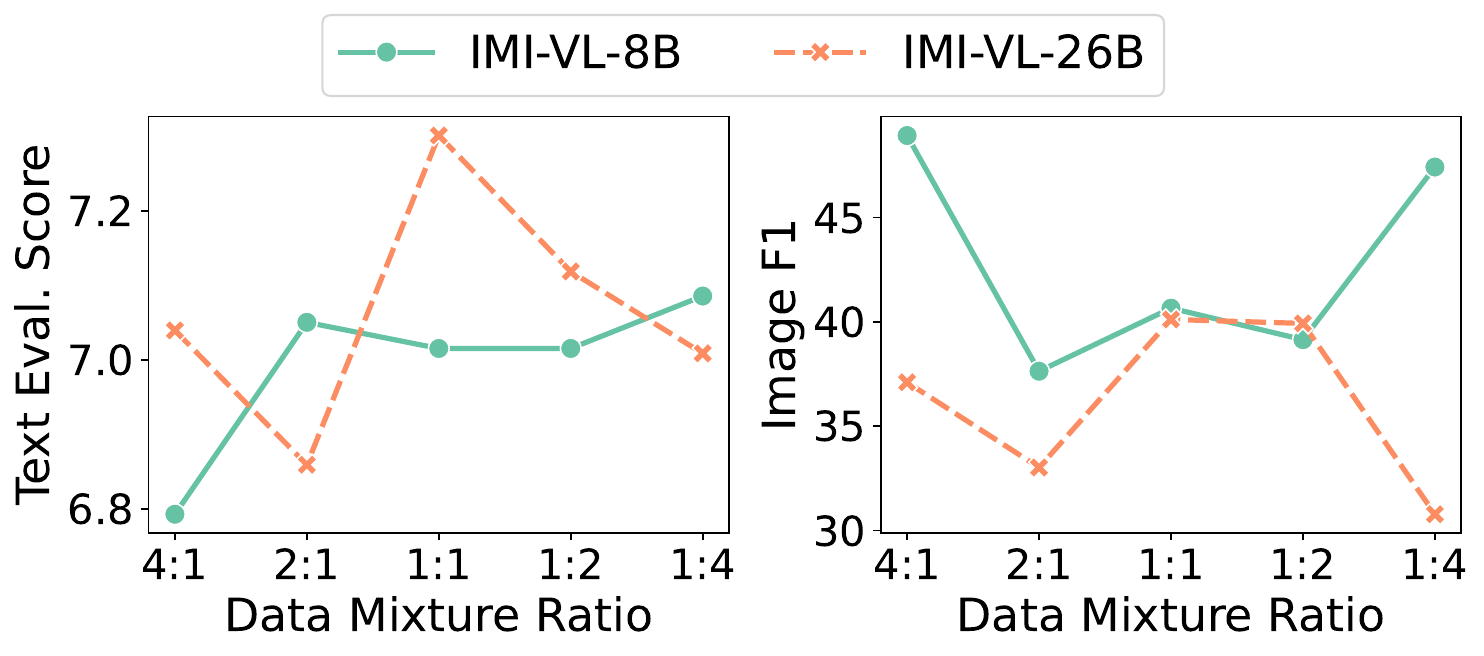}\label{fig:hp-ratio}}
  \subfigure[Training steps.]{\includegraphics[width=0.47\textwidth]{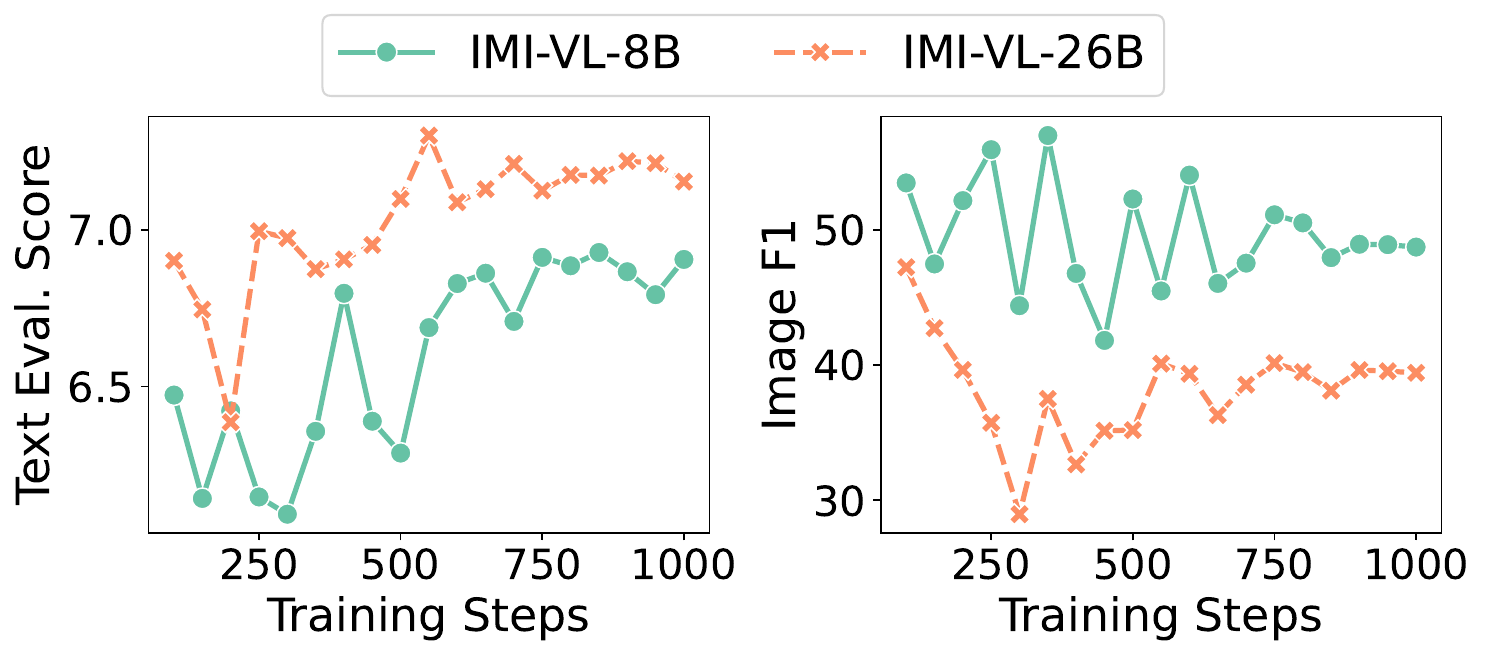}\label{fig:hp-step}}
  \caption{Impact of the hyper-parameters on our \method-8B and \method-26B.}
  \label{fig:hyper}
\end{figure*}

\begin{figure}[!t]
  \centering
  \includegraphics[width=0.40\textwidth]{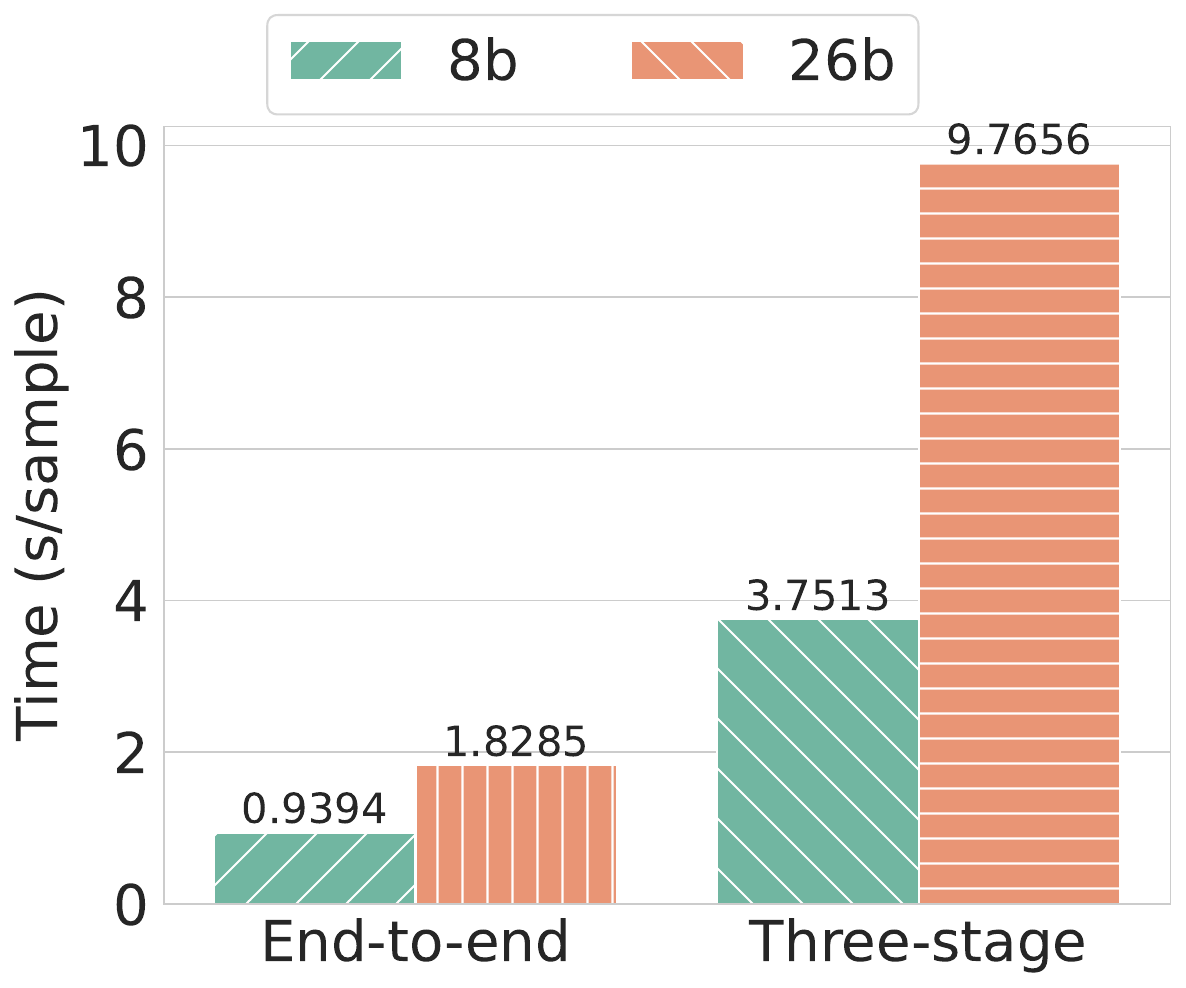}
  \caption{Computational cost comparison between our \method and three-stage methods.}
  \label{fig:speed}
\end{figure}

\noindent \textbf{Open-source VLMs are significantly inferior to that of closed-source VLMs.} Besides, among open-source VLMs, InternVL2 outperforms the others. Currently, open-source VLMs have not considered multi-image contextual understanding tasks during the SFT phase, which could be a reason for their weaker performance. Additionally, there is an inherent performance gap between open-source VLMs and state-of-the-art closed-source VLMs. Among the three open-source VLMs we tested, InternVL2 considered contextual multi-image input during the pretraining phase, which might explain why it performs better.

\noindent \textbf{Three-stage approach yields better results compared to direct end-to-end inference.} The LLMs ability to understand long-context multi-image scenarios is weaker than its ability to perform text-based tasks. Additionally, the caption generation in the three-stage approach benefits from in-context learning, which provides a solid foundation for accurate illustration in the final stage.

\noindent \textbf{Effectiveness of Automatic Metrics.} 
In addition to manual evaluation metrics, we propose two types of automatic evaluation metrics to efficiently conduct preliminary evaluation and model screening. For text evaluation, the Pearson correlation with the text score of human evaluation is 0.9033 ($p < 0.005$), demonstrating its validity. For image position evaluation, due to differences in illustration processes between the three-stage and end-to-end approaches, a fair comparison is not feasible. Excluding the three-stage approach, we calculate the Pearson correlation between the F1 score and image score of human evaluation as 0.9854 ($p < 0.005$), confirming the metric's validity.


\subsection{Computation Overhead Analysis}

In this section, we compare the computational costs of \method, an end-to-end interleaved response generation approach, with a three-stage generation approach. 
Define the number of images in a sample as $N$, the textual context length for each image as $L$, the total context length as $M$, the number of tokens occupied by a single image in the VLM as $P$, the response length as $R$, and 
the length of each image caption is $C$.
The computational complexity of end-to-end method is 
\begin{align}
    O((M + NP + R)^2),
\end{align}
while the complexity of three-stage method is
\begin{equation}
    \begin{aligned}
    & O((M+R)^2 + 9N(L+P+C)^2 \\
    &+ 9(R+NC)^2).
\end{aligned}
\end{equation}
To more intuitively understand the difference in computational costs between the two methods, Figure~\ref{fig:speed} contrasts \method-8B and \method-26B with their respective three-stage InternVL2-8B and InternVL2-26B counterparts.
More specifically, we tested the end-to-end and 3-stage approaches of the 8B and 26B models on a machine equipped with 8 NVIDIA A100-SXM4-80GB GPUs to measure the execution time.
The experimental results demonstrate that the end-to-end approach significantly reduces computational overhead compared to the three-stage scheme.



\subsection{Hyper-parameter Analysis}

Throughout the experiment, we focus on the impact of two hyper-parameters on the results: the proportion of mixed InternVL2 SFT data and the training steps of the model.

\noindent \textbf{Data Mixture Ratio.}
Figure~\ref{fig:hp-ratio} shows the results for different data mixture ratios. For \method-26B, both the text evaluation score and image position evaluation F1 score initially rise and then decline, peaking at around a 1:1 data mixture ratio. In contrast, \method-8B's text evaluation score steadily increases with a higher MI data proportion, while the image position evaluation F1 score first drops and then rises.

\noindent \textbf{Training Steps.}
The results of different data mixture ratio are shown in Figure~\ref{fig:hp-step}.
For \method-26B, the text evaluation score of the model increases with the number of training steps, converging and stabilizing around 500 steps. The F1 score of image position evaluation initially fluctuates but also stabilizes around 500 steps. \method-8B shows a similar trend, but the convergence occurs around 700 steps.


\section{Conclusion}

In this paper, we introduced \textit{Contextual Image Reference} as a novel capability for Vision-Language Models and presented \method, a framework that significantly advances the state-of-the-art in this domain.
Through our carefully curated training data and proposed fine-tuning approach, we demonstrated substantial improvements in VLMs' ability to incorporate relevant images contextually in their responses. 
Our comprehensive evaluation framework, including both automated metrics and human assessment, validates the effectiveness of our approach.
\method demonstrates superior performance over baseline models, achieving significantly better contextual image referencing while being computationally more efficient than multi-stage approaches. 
Our end-to-end system advances the development of visually-aware conversational AI. 
Future work could explore dynamic image generation and modification capabilities. 
We believe this work provides a strong foundation for research in multimodal AI systems with enhanced visual understanding.

\section*{Limitations}
Our work validates the capability of enhanced VLMs to perform interleaved multi-image understanding, thereby exploring the feasibility of contextual image reference. However, our current experiments are based on post-finetuning of InternVL2. Starting from a well-pretrained VLM and incorporating the collected dataset into the supervised finetuning phase might yield better results. Additionally, the current model still exhibits a probability of bad cases. On one hand, collecting more and richer training datasets might address this issue; on the other hand, designing rewards and leveraging techniques like RLHF or RLAIF could be employed for further fine-tuning of the model.

\bibliography{custom}



\end{document}